\documentclass[sigconf]{acmart}
\usepackage[T1]{fontenc}
\usepackage{placeins}

\AtBeginDocument{%
  }

\setcopyright{CC}
\copyrightyear{2023}
\acmYear{2023}
\acmDOI{XXXXXXX.XXXXXXX}

\usepackage[normalem]{ulem}
\usepackage{url}

\DeclareUrlCommand\ULurl{%
  \renewcommand\UrlLeft{\bgroup}%
  \renewcommand\UrlRight{\egroup}}

\acmConference[In2Writing]{Make sure to enter the correct
  conference title from your rights confirmation email}{April 23}{Hamburg, Germany}
\acmBooktitle{In2Writing '23: The Second Workshop on Intelligent and Interactive Writing Assistants
 Apr 23 2023, Hamburg, Germany}

\usepackage{xspace}
\newcommand{\dataset}{\textsc{ManuScript}\xspace}
\usepackage{booktabs}
\usepackage{makecell}
\usepackage{multirow}
\usepackage{adjustbox}

\setcopyright{none}
\renewcommand\footnotetextcopyrightpermission[1]{}
\begin{document}

\title{Decoding the End-to-end Writing Trajectory in Scholarly Manuscripts} 

\author{Ryan Koo}
\authornote{Denotes equal contribution.}
\affiliation{%
  \institution{University of Minnesota}
  \country{Minneapolis, MN, USA}
}

\author{Anna Martin-Boyle}
\authornotemark[1]
\affiliation{%
  \institution{University of Minnesota}
  \country{Minneapolis, MN, USA}
}

\author{Linghe Wang}
\affiliation{%
  \institution{University of Minnesota}
  \country{Minneapolis, MN, USA}
}

\author{Dongyeop Kang}
\affiliation{%
  \institution{University of Minnesota}
  \country{Minneapolis, MN, USA}
}

\renewcommand{\shortauthors}{Koo et al.}

\begin{abstract}
Scholarly writing presents a complex space that generally follows a methodical procedure to plan and produce both rationally sound and creative compositions. Recent works involving large language models (LLM) demonstrate considerable success in text generation and revision tasks; however, LLMs still struggle to provide structural and creative feedback on the document level that is crucial to academic writing. In this paper, we introduce a novel taxonomy that categorizes scholarly writing behaviors according to intention, writer actions, and the information types of the written data. We also provide \dataset, an original dataset annotated with a simplified version of our taxonomy to show writer actions and the intentions behind them. Motivated by cognitive writing theory, our taxonomy for scientific papers includes three levels of categorization in order to trace the general writing flow and identify the distinct writer activities embedded within each higher-level process. \dataset intends to provide a complete picture of the scholarly writing process by capturing the linearity and non-linearity of writing trajectory, such that writing assistants can provide stronger feedback and suggestions on an end-to-end level. 
The collected writing trajectories are viewed at \href{https://minnesotanlp.github.io/REWARD_demo/}{\color{blue}{\url{https://minnesotanlp.github.io/REWARD\_demo/}}}\footnote{
The public code for the data collection and Chrome extension is here: 
\href{https://github.com/minnesotanlp/reward-system}{\color{blue}{\url{https://github.com/minnesotanlp/reward-system}}}}
\end{abstract}




\maketitle

\section{Introduction}
Writing is a cognitively active task involving continuous decision-making, heavy use of working memory, and frequent switching between multiple activities. Scholarly writing is particularly complex as it requires the author to coordinate many pieces of multiform information while also meeting the high standards of academic communication. Flower and Hayes' \cite{flower-hayes-1981} cognitive process theory of writing organizes these tasks into three processes: \textit{planning}, during which the writer generates and organizes ideas and sets writing goals; \textit{translation}, during which the writer implements their plan, keeping in mind the organization of the text as well as word choice and phrasing; and \textit{reviewing}, during which the writer evaluates and revises their text. Flower and Hayes emphasize that these distinct phases are non-linear and highly embedded, meaning that any process or sub-process can be embedded within any other process and move back and forth between each process. In order to provide relevant feedback at each step of the academic writing process, it is critical for writing assistants to have a strong understanding of the planning, translation, and revision stages throughout their entirety.

\begin{figure}[t]
    \centering
     \includegraphics[width=0.5\textwidth,trim={0cm 6cm 9cm 3cm},clip]{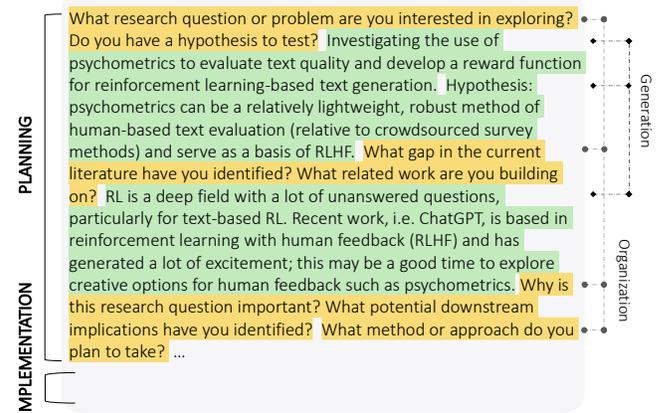}
      \vspace{-6mm}
    \caption{An example manuscript with annotations on writing intentions (left) and writing actions (right). Each horizontal line denotes a single annotation.}
    \label{fig:annotation_example}
    \vspace{-5mm}
\end{figure}

Recent corpora for the study of writing processes exist for each of these sub-processes. Berdanier \cite{berdanier_2016} demystifies the academic writing process in a study showing the ``linguistic scheme'' involving a distinct planning and crafting procedure typically followed within technical writing. Furthermore, much work has been done to study text revision using keystroke data \cite{conijn-etal-2020-process,https://doi.org/10.1002/ets2.12247,Ameri2017WritersOT}, and revision history \cite{du-etal-2022-understanding-iterative,zhang-etal-2017-corpus,yang-etal-2017-identifying-semantic,ito-etal-2019-diamonds,daxenberger-gurevych-2012-corpus}. More recently, Sardo et al. \cite{Sardo2023ExploitationAE} have developed a corpus and a metric for edit-complexity that draws a complex topological structure of the writer's efforts throughout the history of the essay to study the planning and translation processes. Despite recent advancements in large language models, particularly text generation, LLMs still exhibit subpar performance for reasoning capabilities and particularly planning \cite{valmeekam-2022} to have any significant impact in aiding the writing process \cite{Sardo2023ExploitationAE}. Our work builds upon these previous studies to provide a dataset with annotations encompassing the writing process spanning across all three stages, as described by Flower and Hayes.

Our contributions include \dataset, a small dataset of scholarly writing actions, and a comprehensive taxonomy of writing processes that are applicable across various academic disciplines. \dataset is annotated following a simplified version of our taxonomy to capture the end-to-end writing process. Our work is motivated by the idea that providing writing assistants  detailed information about the writing process will help them give more appropriate suggestions to writers throughout the writing process. Applying this taxonomy to a dataset of academic writing samples will give us insight into the academic writing process and provide us with data to support the generation of suggestions that align with the writer's current activity and intention. In the future, we plan to extend this work by scaling the data collection process over a longer period of time to develop a more nuanced taxonomy of writers' actions and intentions.

\begin{figure}[tbp!]
    \centering
    \includegraphics[width=0.5\textwidth,trim={0 3.5cm 0 0},clip]{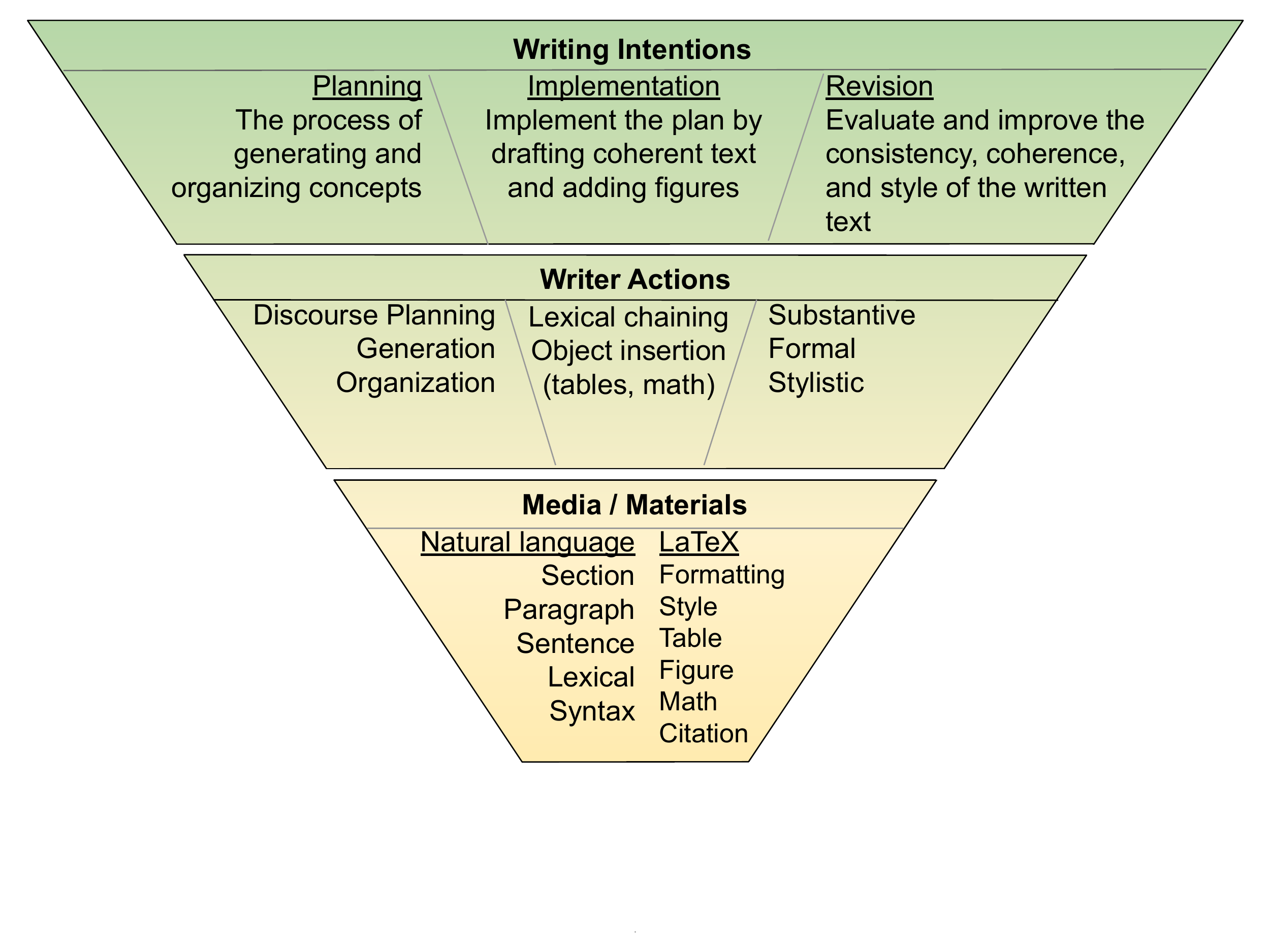}
    \caption{Hierarchical Taxonomy of Writing Actions}
    \label{fig:taxonomy}
    \vspace{-1.2cm}
\end{figure}
\section{\dataset: A Dataset of the End-to-End Writing Process}

Analyzing a final manuscript alone is intractable for capturing an author's original intentions.
We have developed a taxonomy of scholarly writing trajectories illustrated by Figure \ref{fig:taxonomy} that can characterize the finer-grained actions an author takes into distinct categories but is also general enough to fully capture the author's trajectory throughout the entire writing process. The highest level of our taxonomy describes the intention informing the writer's actions, and is based on the three main processes described by Flower and Hayes \cite{flower-hayes-1981}. The middle layer describes the various writing actions that might take place to carry out the writer's intention. Each intention is associated with its own set of actions. For example, while the author is revising their work, they may be making substantive, formal, or stylistic revisions. The lowest level describes the linguistic or LaTeX unit that they are currently operating on. For example, if the writer is drafting and moving around paragraph topic sentences within a new section of their paper, their spans of keystrokes would alternate between \textsc{Planning $\to$ Generation $\to$ Section} and \textsc{Planning $\to$ Organization $\to$ Section} because they are working at the section level and switching between generating new ideas and organizing them.

\paragraph{Data Collection}
We developed a chrome extension that reverse engineers Overleaf's editing history utilizing user keystrokes to track writing actions in real-time (See details in Appendix \ref{appendix:tracking}).
From this, we can generate a playback that shows the chronological progression for each completed writing session. Our initial study involved four participants in a pilot study where they were prompted to describe their current or future research plans by responding to the available prompts or in free form over a thirty-minute writing session.  \begin{table}[H]
    \centering
    \begin{tabular}{@{}p{1.6cm} p{6cm}}
        \textbf{Label} & \textbf{Description} \\ \toprule
        \multicolumn{2}{@{}p{8cm}@{}}{\textbf{\textsc{Planning}}  The writer's intention is to get their ideas down on paper in a semi-structured manner.}
        \\ 
        \cmidrule{1-2}
        \textit{generation} & The process of adding ideas to the document.\\
        \cmidrule{2-2}
        \textit{organization} & Structuring the generated concepts. \\ 
        \midrule
        \multicolumn{2}{@{}p{8cm}@{}}{\textbf{\textsc{Implementation}}  The writer's intention is to produce high-quality and persuasive text that meets their writing goals.}
        \\
        \cmidrule{1-2}
        \multirow{3}{*}{\makecell{\textit{lexical}\\  \textit{chaining}}} & Writing coherent text where sentences are linked by the semantic relationships between words \cite{morris1991}. \\  
        \midrule
        \multicolumn{2}{@{}p{8cm}@{}}{\textbf{\textsc{Revision}} The writer's intention is to improve the clarity, consistency, coherence, and style of the written text.}\\ 
        \cmidrule{1-2}
        \textit{syntactic} & Fixing grammar, spelling, and punctuation.\\
        \cmidrule{2-2}
        \textit{lexical} & Changing words to clarify meaning or improve coherence.\\
        \cmidrule{2-2}
        \textit{structural} & Reordering text to improve organization.\\ 
        \bottomrule
    \end{tabular}
    \caption{Simplified annotation schema applied to our dataset}
    \label{tab:schema}
     \vspace{-8mm}
\end{table}
 In total, we collected four writing trajectories, including 46 discontinuous edits with 3290 recorded actions. The detailed statistics are in Appendix \ref{appendix:statistics}.

\paragraph{Annotation Schema}

Due to the limited scope of our pilot study, we applied a reduced annotation schema, containing two levels of granularity (Table \ref{tab:schema}). 
The higher level includes \textsc{Planning}, \textsc{Implementation}, and \textsc{Revision}. These labels are used to denote the general process that the writer is working in. 
The lower level categorizations include $\{$idea \textsc{generation}, concept \textsc{organization}$\}$, $\{$\textsc{lexical$\_$chaining}$\}$, and $\{$\textsc{syntactic, lexical, structural}$\}$ for each of the three processes respectively. Presently, the category of \textsc{Implementation} is limited in that the only sub-category is \textsc{lexical$\_$chaining}. We hope to learn more about the \textsc{Implementation} process during our next study.
\begin{figure}[h]
    \centering
    \includegraphics[width=0.5\textwidth,trim={0cm 16.4cm 17cm 3cm},clip]{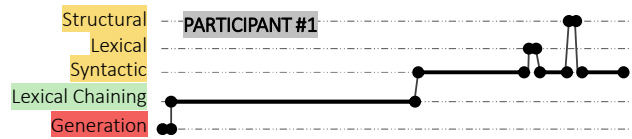}
     \vspace{-6mm}
    \caption{Annotated writing trajectory of one participant. The x-axis shows the writing steps chronologically. The horizontal bands show the three high-level processes of Planning, Implementation, and Revision.}
    \label{fig:timeseries_selected}
    \vspace{-4mm}
\end{figure}

\section{Annotation Results}

Three of the authors annotated the samples that were gathered (See Figure \ref{fig:annotation_example} for an example). One author annotated sample 1 in the course of developing the annotation guidelines. Figure \ref{fig:timeseries_selected} illustrates the first participant's writing trajectory. Each of the other three samples was annotated by two different authors such that each author annotated two samples, and no two samples had the same pair of annotators.
The inter-annotator agreement score (mean F1) across the three samples is 65.26.
For all scores, see Appendix \ref{appendix:iaa}.

\section{Future Work}
\paragraph{Extended schema} The simplified annotation schema we applied to our data is limited in its ability to capture the expressiveness and nuance of scholarly communication. To this end, we are continuing to refine the hierarchical taxonomy of scholarly writing (see Figure \ref{fig:taxonomy}). 
For example, while revising their work, a writer might replace a word with another to improve clarity; this would be classified as \textsc{Revision}$\to$\textsc{Substantive}$\to$\textsc{Lexical}.

\paragraph{Larger data collection} 
To validate our taxonomy and gain deeper insight into the scholarly writing process, we will need to collect more writing data over a longer period of time. The current study design is too short (30 minutes), and the prompt is too limiting to gather a comprehensive representation of scholarly writing behaviors. Our future study will be conducted over a period of months and will observe the writing actions of researchers as they write their actual academic works in order to elicit data that accurately represents the scholarly writing process.

\paragraph{With multiple authors}
Often within the writing process for scholarly papers, multiple authors will work on a manuscript simultaneously. For example, the input of other authors, comments, or suggestions may influence an author's writing trajectory compared to their usual writing habits in an individual setting. Therefore, tracking how the writing trajectory differs between the individual writing space and the collaborative one poses an interesting task to explore. 

\paragraph{Multiple academic disciplines.} The authors of this work have a bias towards writing conventions in computer science research. While we developed our taxonomy to be general enough to be applied to various academic disciplines, there may be nuances in the writing requirements for other disciplines that we are unfamiliar with. Further study is required to ascertain appropriate modifications to our schema for different disciplines. In particular, we believe the writer actions that belong to the \textsc{Implementation} phase might need to be expanded for other disciplines, and additional information units added to the \textsc{Media/Materials} level.

\paragraph{Writing Assistants} \dataset intends to decode the writing process in academic writing by capturing writer actions in an end-to-end manner such that writing assistants can provide more useful feedback at each phase of the process. Through taxonomizing writer actions at each point, the dataset can provide a good representation of the trajectory that authors tend to take within their writing and their intentions that may provide current writing assistants with a more clear understanding in predicting the next steps that the writer envisions.

\newpage
\bibliographystyle{ACM-Reference-Format}
\bibliography{custom}

\clearpage
\appendix
\section{Writing Action Tracking System} \label{appendix:tracking}
Since a single-character record does not provide any useful information about a user's writing actions and intentions, we process each character level by grouping them to form word- and sentence-level actions to extract comprehensible edits that paint a more meaningful picture of their writing topography. First, each time the user types a space, enters a carriage return, leaves the tab, copies/pastes/cuts, or switches files, the text currently seen by the user is recorded. Then, we utilize the \texttt{diff\_match\_patch} \footnote{\url{https://github.com/google/diff-match-patch}} library to extract the differences between the last and current recorded content to find the most recent edit. 

\begin{table}[H]
    \centering
    \begin{tabular}{cccc}
        \textbf{Sample} & \textbf{F} & \textbf{P} & \textbf{R} \\ \toprule
        2 & 00.8 & 00.8 & 00.8 \\ 
        3 & 96.6 & 96.9 & 96.4 \\
        4 & 98.4 & 98.0 & 98.79 \\ \midrule
        Mean & 65.26 & 65.20 & 65.20 \\ \bottomrule
    \end{tabular}
    \caption{Inter-annotator agreement F1, Precision, and Recall scores for each sample.}
    \label{tab:iaa-results}
\end{table}
\section{Annotation Scores} \label{appendix:iaa}
Inter-Annotator Agreement was measured by calculating the F1, Precision, and Recall scores in a multi-label, multi-class setting (see Table \ref{tab:iaa-results} for the results). To prepare a pair of annotations for scoring, each unit of text for each sample was treated as a slot containing a ten-digit bitmap, where each bit represents a different label. 
Note that sample two had a near-zero agreement between the annotators. This occurred because of the similarity between the Planning activity of idea \textsc{generation} and the Implementation activity of \textsc{lexical\_chaining}. Sample two was markedly different from all other samples in that the participant composed the entire sample linearly from start to finish in perfect, coherent English without going back to change anything or doing any initial document planning. The guidelines were ambiguous for this sample. One annotator marked this text as \textsc{generation} since the participant started drafting from scratch. The other annotator labeled this sample as \textsc{Implementation}, since the participant was creating fully-formed paragraphs that could appear in the final draft. 

This suggests that the annotator sometimes has to see into the future of the document in order to annotate confidently. If participant two continued working on this document for another few hours, we could tell whether these first steps were Planning or Implementation. If they had gone back and expanded on each of the paragraphs they drafted, then it would be clear that the first steps were a Planning process. If they continued to draft this way until they were done writing the document, then it would be clear that these first steps were an Implementation process. In this case, we would assume that the Planning process happened solely in his head or in an external document. A future study should have an audio component where the participant narrates their process to provide insight into the writing intentions. Furthermore, we observe that participant 2 wrote the way a student may write during a timed essay examination. Future study design should give participants more time to work on their sample, perhaps extending over several sessions.
\section{Data and Annotation Statistics}\label{appendix:statistics}
Sample 1 exhibited the most additions/deletions, with Sample 2 showing the second most additions and the fewest deletions in Table \ref{tab:data_statistics} but had the highest lexical-chaining value in table 4. Therefore, Sample 2 writers spent most of their time writing paragraphs. Sample 3 has the middle number of added and deleted words, with the highest "generation" and "organization" in Table \ref{tab:annotation_statistics}, indicating that most of the content is planning. We can also infer that the number of words planned is less than the number in formal writing. Sample 4 has the lowest number of words added. Similarly to sample 3, both annotators classify sample 4  entirely as "planning," but generation and organization are smaller than in sample 3, which explains why there were fewer words added, as seen in Table \ref{tab:data_statistics}.
\begin{table}[h]
\resizebox{0.5\textwidth}{!}{%
\begin{tabular}{ p{1cm}p{1.5cm}p{1.3cm}p{1.3cm}p{1.3cm}}
 \toprule
Sample &No. \newline disc-edits &Added words &Deleted words &Recorded actions\\
 \midrule
 1 &11  &1304   &348    &1167\\
 2 &4   &886    &13     &808\\
 3 &23  &769    &39     &687\\
 4 &9   &692    &52     &628\\
 \bottomrule
\end{tabular}}
 \caption{The numbers of discontinuous edits, added and deleted words, and total actions per sample of the   \dataset dataset.}
 \label{tab:data_statistics}
 \end{table}
\begin{table}[H]
  \centering
  \resizebox{0.35\textwidth}{!}{%
  \begin{tabular}{@{}lcccc@{}}
    & \multicolumn{4}{c}{Samples} \\
    \cmidrule(l){1-5} & 1 & 2 & 3 & 4 \\
    \cmidrule(l){1-5}
    \textsc{Planning} &  1.0 &  0.5 &  1.0 &  1.0 \\
    \cmidrule(l){1-1}\cmidrule(l){2-5}
    \hspace{5mm} \textsc{Generation} & 1.0 & 0.0 & 15.0 & 5.5 \\
    \hspace{5mm} \textsc{Organization} & 1.0 & 0.5 & 10.0 & 4.5 \\
    \cmidrule(l){1-1}\cmidrule(l){2-5}
    \textsc{Implementation} & 3.0 & 0.5 & 0.0 & 0.0 \\
    \cmidrule(l){1-1}\cmidrule(l){2-5}
    \hspace{5mm} \textsc{lex$\_$chaining} & 3.0 & 3.0 & 0.0 & 0.0 \\
    \cmidrule(l){1-1}\cmidrule(l){2-5}
    \textsc{Revision} & 2.0 & 0.0 & 0.0 & 0.0 \\
    \cmidrule(l){1-1}\cmidrule(l){2-5}
    \hspace{5mm}\textsc{Syntactic} & 0.0 & 0 & 0.0 & 0.0 \\
    \hspace{5mm}\textsc{Lexical} & 1.0 & 0.0 & 0.0 & 0.0 \\
    \hspace{5mm}\textsc{Structural} & 1.0 & 0.0 & 0.0 & 0.0 \\
    \cmidrule(l){1-1}\cmidrule(l){2-5}
    \textsc{None} & 1.0 & 0.5  & 0.5 & 0.5 \\
    \bottomrule
  \end{tabular}}
 \caption{The distribution of labels per sample (averaged over 2 annotators)}
 \label{tab:annotation_statistics}
\end{table}

\begin{figure}[H]
    \centering
    \includegraphics[width=0.5\textwidth]{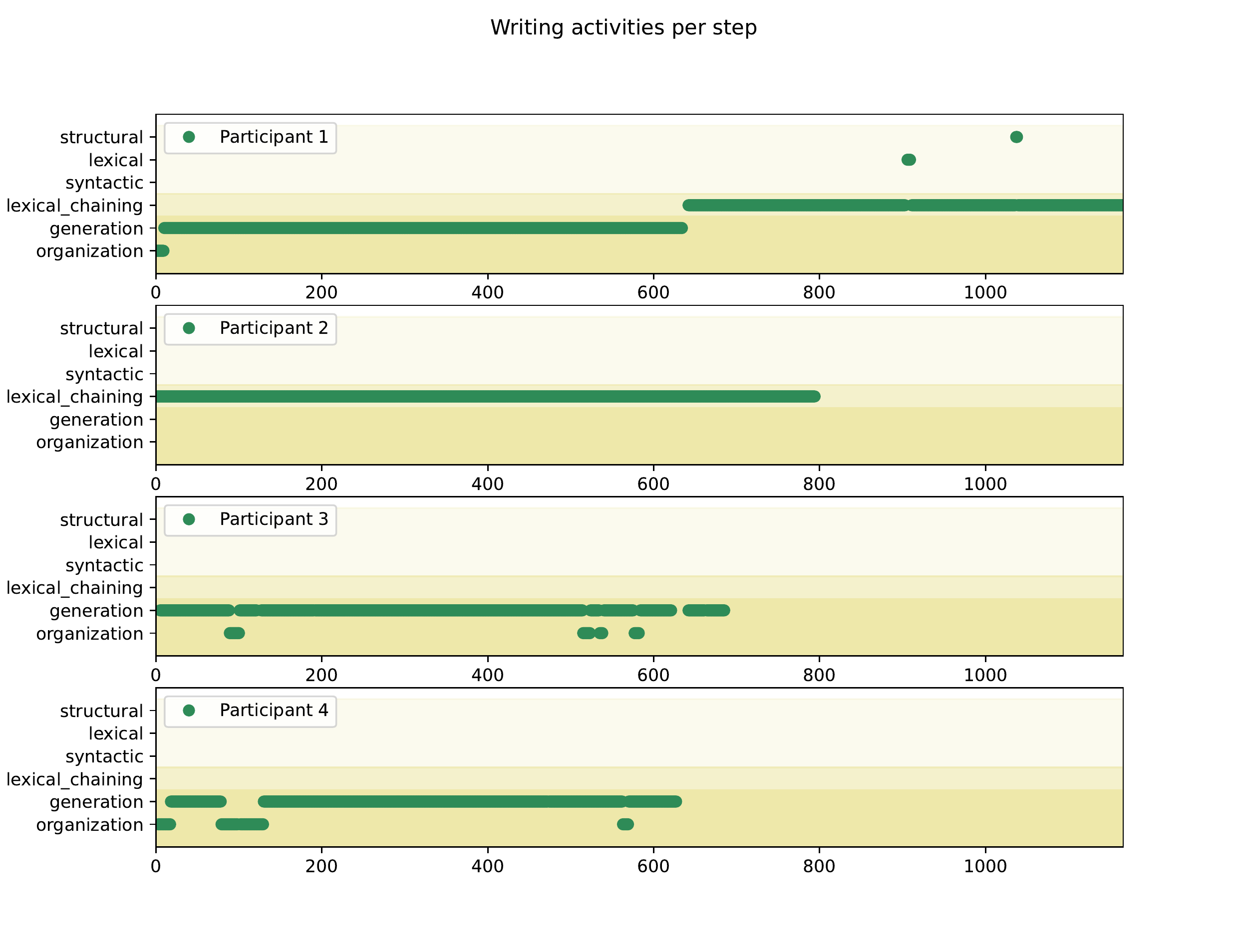}
    \caption{This shows the label assigned to each writing step that each participant wrote. The x-axis shows the writing steps chronologically. The horizontal bands show the three high-level processes of Planning (bottom), Implementation (middle), and Revision (top).}
    \label{fig:timeseries}
\end{figure}
Figure \ref{fig:timeseries} shows the participants' actions in chronological order throughout the study session. Notice that the entirety of the study is spent in the Planning phase for participants three and four. Participant one spends a similar amount of actions in the Planning phase as participants three and four, but editing more quickly, was able to move into Implementation and even Revision phases towards the end. Participant two is an outlier; likely, they are implementing an internal plan rather than planning in the document first.

\newpage
\section{Annotation Schema and Taxonomy Design}
\paragraph{Simple Schema}To identify the writer's intentions at each point, we categorize each higher-level span into various lower-level ones specific to the different processes. 
The \textsc{Planning} process involves the point in which the writer starts generating and organizing concepts and arguments, such as drafting topic sentences or simple paragraphs, and could also take the form of more fragmented language. \textsc{Planning} can be branched into idea \textsc{generation} where the writer gets their ideas down on the page and concept \textsc{organization} where the writer is structuring their concepts, arguments, and topics. The \textsc{Implementation} process can be described as when the author starts implementing their plan by drafting full sentences and paragraphs, potentially rewriting material from the Planning process to fit in with the full context they are generating. We break this down into distinct periods of \textsc{lexical\_chaining} in which a sequence of sentences are linked by the semantic relationships between the words in the sentences \cite{morris1991}. The \textsc{Revision} process can be broken down into \textsc{syntactic}revisions, \textsc{lexical} revisions, and \textsc{structural} revisions. The label \textsc{None} is used when no other label is suitable. 

\paragraph{Extended Schema} While we used the simple schema described above to annotate our preliminary results, we intend to apply a more complex schema to future studies. To support a more complex schema, we are developing the taxonomy described in Figure \ref{fig:taxonomy}.

\end{document}